\documentclass[10pt,letterpaper]{article}

\usepackage{times}
\usepackage{url}
\usepackage[hidelinks]{hyperref}
\usepackage[utf8]{inputenc}
\usepackage[small]{caption}
\usepackage{graphicx}
\usepackage{amsmath}
\usepackage{amssymb}
\usepackage{amsthm}
\usepackage{booktabs}
\usepackage{algorithm}
\usepackage{algorithmic}

\providecommand{\citeauthoryear}[2]{#1, #2}

\title{WASHH: An Anchor-Aware Whale-Guided Selection Hyper-Heuristic for Continuous Optimization and SVC Configuration}

\author{
Yifu Zhao$^1$, Xiaofan Zou$^2$, Junhao Wei$^{1,3}$, Yanxiao Li$^1$, Baili Lu$^1$, Zhenhong Peng$^4$\\
Dexing Yao$^1$, Haochen Li$^1$, Qinbin He$^1$, Sio-Kei Im$^5$, Xu Yang$^1$, Yapeng Wang$^1$\thanks{Corresponding author: yapengwang@mpu.edu.mo}\\[0.5em]
\small $^1$Faculty of Applied Sciences, Macao Polytechnic University, Macao, 999078, China\\
\small $^2$School of Mechanical and Electrical Engineering and Automation, Shanghai University, Shanghai, 200444, China\\
\small $^3$Pazhou Lab (Huangpu), Guangzhou, 510555, China\\
\small $^4$College of Animal Science and Technology, Zhongkai University of Agriculture and Engineering, Guangzhou, 510225, China\\
\small $^5$Macao Polytechnic University, Macao, 999078, China
}

\date{}

\begin{document}

\maketitle

\begin{abstract}
Learning-assisted algorithm design often has to make reliable search decisions under small evaluation budgets, where committing to a single metaheuristic can be unreliable.
We propose WASHH, a Whale-guided Adaptive Selection Hyper-Heuristic for continuous black-box optimization.
WASHH uses WOA as the main exploitation backbone, but treats PSO-style memory, GWO-style leader averaging, DE-style variation, local coordinate search, and anchor-guided refinement as selectable search behaviors.
An online reward controller allocates evaluations according to observed improvements, while anchor refinement exploits inexpensive reference configurations such as box centers or default model settings without bypassing black-box evaluation.
On ten 30-dimensional benchmark functions with 10 independent runs and 12,000 evaluations, WASHH achieves the best average rank, 1.10, and is best or tied best on all ten functions.
It strictly improves over WOA on eight functions and ties WOA at the numerical optimum on Rastrigin and Griewank.
We further study SVC hyperparameter configuration for breast cancer diagnosis under a 300-evaluation budget.
WASHH obtains the lowest mean validation log loss among the compared optimizers, suggesting that anchor-aware selection hyper-heuristics are a practical lightweight direction for LEAD systems.
\end{abstract}

\section{Introduction}

Continuous black-box optimization appears in algorithm configuration, simulation-based design, hyperparameter optimization, and data-driven evolutionary systems.
In these settings, the landscape is usually unknown before the evaluation budget is spent, and the search behavior that works well in one phase may be ineffective in another.
Exploitative attraction toward the incumbent can accelerate convergence on smooth basins, whereas memory, multi-leader guidance, differential variation, and local refinement provide different ways to balance stability and diversity.
Hyper-heuristics are therefore attractive because they decide how to use lower-level search heuristics during the run rather than forcing the entire optimization process to follow one fixed rule~\cite{cowling2001hyperheuristics,burke2013hyper}.

This question is closely aligned with the LEAD setting.
A learning-assisted optimizer can be built from reusable operators, but a new problem instance may allow only a small number of expensive evaluations.
When offline training data are limited or unavailable, the optimizer must adapt from the improvements observed online.
The challenge is to obtain this adaptivity without adding a heavy learning component that consumes the budget it is meant to save.

We propose WASHH, a Whale-guided Adaptive Selection Hyper-Heuristic.
WOA~\cite{mirjalili2016woa} provides a strong exploitation geometry, while the controller can also select PSO-style memory~\cite{kennedy1995particle}, GWO-style multi-leader averaging~\cite{mirjalili2014grey}, DE-style differential variation~\cite{storn1997differential}, local coordinate refinement, and anchor-based refinement.
The anchor component captures a common feature of algorithm-configuration tasks: low-cost reference points often exist, including default hyperparameters, box centers, or historically effective configurations.
WASHH uses these references to guide refinement, but every proposed solution is still judged by the original black-box objective.

The paper contributes an adaptive selection hyper-heuristic that combines WOA-guided search, online behavior selection, and anchor refinement; a reproducible comparison with WOA, GWO, PSO, DE, LWOA, and RandomHH on ten continuous benchmark functions; and a breast cancer diagnosis case study in which WASHH configures an SVC model under a fixed evaluation budget.

\begin{figure*}[t]
\centering
\includegraphics[width=0.92\textwidth]{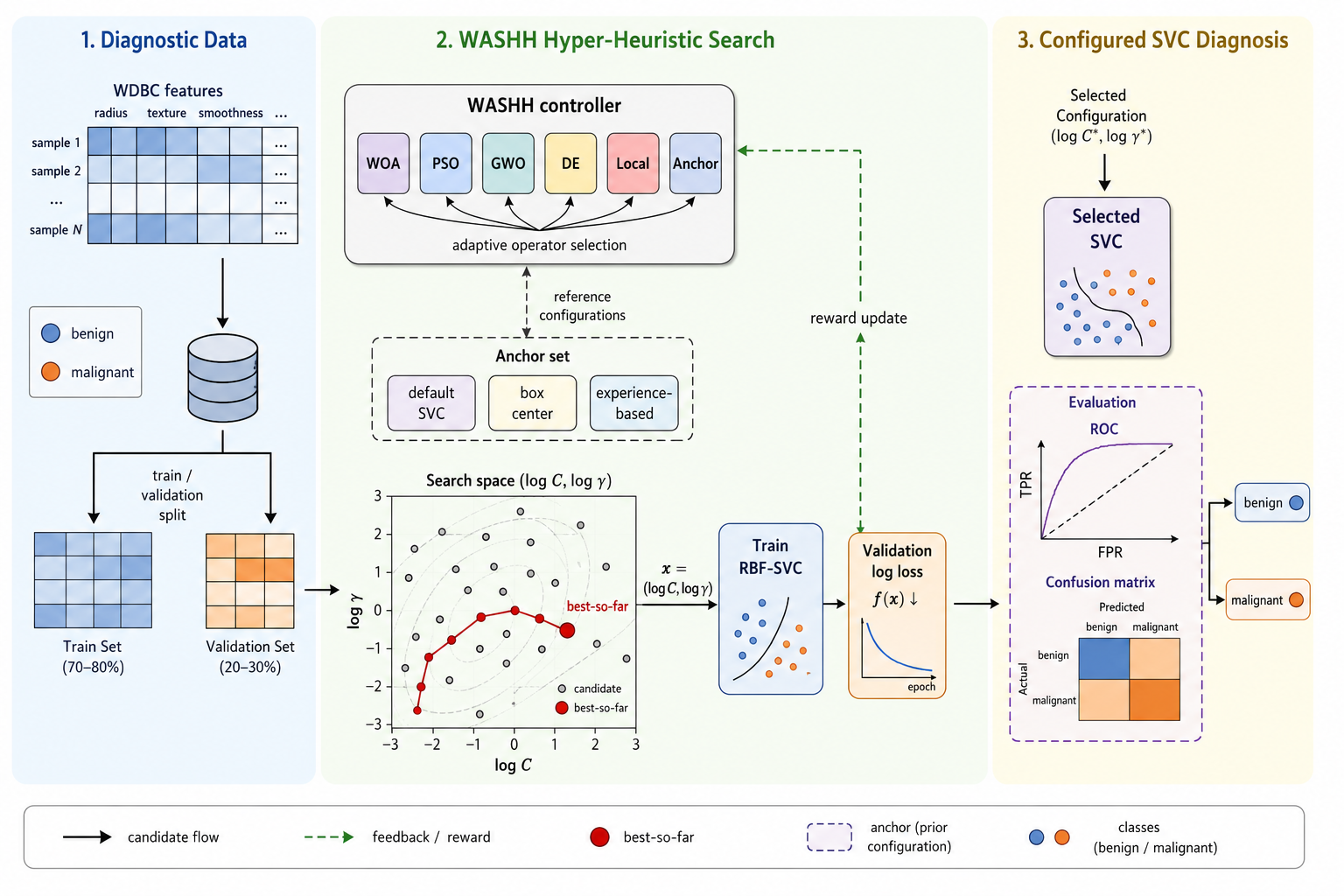}
\caption{Schematic overview of breast cancer diagnosis model configuration with WASHH. Candidate SVC hyperparameters are generated by the adaptive hyper-heuristic controller, evaluated through validation log loss, and returned as feedback for subsequent operator selection and anchor-guided refinement.}
\label{fig:framework}
\end{figure*}

\section{Related Work}

Hyper-heuristics provide a natural framework for LEAD because they separate the design of search behaviors from the decision of when to use them.
Instead of assuming that one low-level heuristic should control the whole run, a selection hyper-heuristic monitors the search process and reallocates trials among available behaviors~\cite{cowling2001hyperheuristics,burke2013hyper}.
This view is especially relevant under small evaluation budgets: offline landscape models may be unavailable, but recent improvements can still be converted into online rewards for operator choice~\cite{fialho2010analyzing}.
WASHH follows this budget-conscious setting by adapting the use of population-search behaviors within a single optimization run.

The operator library used by WASHH is motivated by the complementary biases of established population-based optimizers.
Methods such as WOA~\cite{mirjalili2016woa}, PSO~\cite{kennedy1995particle}, GWO~\cite{mirjalili2014grey}, and DE~\cite{storn1997differential} encode different assumptions about exploitation, memory, leadership, and population variation.
Recent swarm-optimization studies have further shown that carefully designed movement and update mechanisms can improve performance in engineering design, path planning, and prediction tasks~\cite{wei2024adaptive,wei2025tswoa,wei2026geometric,gu2025gwoa,wei2026nawoa,li2026askssa}.
However, most such methods still commit to a fixed update schedule once the algorithm is chosen.
WASHH takes a different position: these mechanisms are treated as reusable search behaviors inside an adaptive controller rather than as mutually exclusive standalone optimizers.

A second motivation comes from algorithm configuration and hyperparameter optimization, where reference configurations are often available before the search begins.
Default parameters, box centers, and experience-based settings are not guaranteed to be optimal, but they often identify regions worth validating.
Anchor refinement uses this information as a search bias while preserving the black-box evaluation loop.
This distinction is important for LEAD systems: prior knowledge can guide where the optimizer searches, but the current task still determines which candidate configurations are accepted.

\section{WASHH}

We consider minimization of $f:\mathbb{R}^d\rightarrow\mathbb{R}$ under bound constraints and a fixed budget $B$.
The population size is $N$.
All candidate evaluations, including initialization, operator trials, and anchor refinements, count toward the same budget.

\subsection{Operator Portfolio}

WASHH represents the search state with a population $P=\{x_i\}_{i=1}^N$, the associated fitness values, the global incumbent $x^\star$, optional personal memories for memory-based moves, and a reward score attached to each low-level behavior.
The behavior portfolio is deliberately heterogeneous.
Its WOA-guided moves provide incumbent-centered encircling and spiral search, memory-based moves reuse personal and global information, differential moves introduce population variation, and multi-leader moves stabilize the update direction with the best three individuals.
These exploratory and exploitative behaviors are complemented by local coordinate refinement and anchor-based refinement, both of which are intended to improve the incumbent once a promising region has been identified.

During optimization, the controller samples a behavior according to normalized reward scores and uses it to propose one candidate solution.
The candidate is clipped to the feasible box and evaluated by the original black-box objective.
Improvements to the selected population slot or to the global incumbent increase the reward of the behavior that generated the candidate.
The scores are then exponentially smoothed, with a small exploration floor that prevents any behavior from disappearing permanently.

\subsection{Anchor-Guided Refinement}

Anchor refinement uses a set $\mathcal{A}$ of reference configurations as search guides rather than as fixed solutions.
On continuous benchmark functions, these references are deterministic points derived from the search domain, including the box center and feasible zero or unit vectors; for wide symmetric domains, the implementation also uses a positive-side reference.
In the breast cancer diagnosis application, the same role is played by default or experience-based SVC hyperparameters.
The anchor proposal has the form
\begin{equation}
x' = a + \gamma(x^\star-a) + \epsilon,
\end{equation}
where $a\in\mathcal{A}$, $\gamma\in[0,1]$, and $\epsilon$ is a small local perturbation.
Near the end of the budget, WASHH reserves a small refinement budget for deterministic anchor-neighborhood and coordinate searches.
The anchor therefore suggests where to intensify the search, while the objective evaluation still decides whether the resulting candidate should be accepted.

\begin{algorithm}[t]
\caption{WASHH}
\label{alg:washh}
\begin{algorithmic}[1]
\STATE Initialize a population using random samples and available anchors
\STATE Evaluate the population and initialize operator scores
\WHILE{main search budget remains}
    \STATE Select an operator according to normalized reward scores
    \STATE Generate a candidate using WOA, PSO, DE, GWO, local, or anchor behavior
    \STATE Clip the candidate to the feasible box and evaluate it
    \STATE Update the population, incumbent, memories, and operator score
\ENDWHILE
\WHILE{reserved refinement budget remains}
    \STATE Evaluate deterministic anchor-neighborhood and coordinate candidates
    \STATE Update the incumbent when an improvement is found
\ENDWHILE
\RETURN best solution found
\end{algorithmic}
\end{algorithm}

\section{Experiments}

\subsection{Benchmark Protocol}

The benchmark suite contains ten 30-dimensional functions: Sphere, Bent Cigar, Zakharov, Rosenbrock, Rastrigin, Ackley, Griewank, Schwefel, Levy, and Michalewicz.
Together, they cover smooth unimodal, ill-conditioned, nonseparable, valley-shaped, multimodal, and deceptive landscapes.
All methods are evaluated with population size $N=30$ and a budget of 12,000 objective evaluations, using 10 independent seeds for each function.
We compare WASHH with WOA, GWO, PSO, DE, LWOA, and RandomHH.
Here LWOA serves as a WOA variant with long-jump exploration, and RandomHH represents a non-adaptive hyper-heuristic that samples uniformly from a heterogeneous operator portfolio.
Performance is reported as the mean final objective value with standard deviation, and average ranks are computed with exact ties on the mean values.

\begin{table*}[t]
\centering
\caption{Average and standard deviation of final objective values over 10 independent runs.}
\label{tab:results}
\setlength{\tabcolsep}{4pt}
\resizebox{\textwidth}{!}{%
\begin{tabular}{llccccccc}
\toprule
Function & Metric & WASHH & WOA & GWO & PSO & DE & LWOA & RandomHH \\
\midrule
Sphere & Ave & $\mathbf{0.0000E{+}00}$ & $1.1722\mathrm{E}{-}62$ & $6.6395\mathrm{E}{-}24$ & $1.3218\mathrm{E}{+}00$ & $3.9040\mathrm{E}{+}00$ & $2.0606\mathrm{E}{-}04$ & $4.3850\mathrm{E}{-}21$ \\
 & Std & $\mathbf{0.0000E{+}00}$ & $3.6988\mathrm{E}{-}62$ & $8.0683\mathrm{E}{-}24$ & $2.0393\mathrm{E}{+}00$ & $7.9522\mathrm{E}{+}00$ & $2.3443\mathrm{E}{-}04$ & $9.5194\mathrm{E}{-}21$ \\
Bent Cigar & Ave & $\mathbf{0.0000E{+}00}$ & $1.8263\mathrm{E}{-}59$ & $4.8310\mathrm{E}{-}18$ & $2.1705\mathrm{E}{+}05$ & $3.9316\mathrm{E}{+}06$ & $2.4312\mathrm{E}{+}02$ & $3.6848\mathrm{E}{-}15$ \\
 & Std & $\mathbf{0.0000E{+}00}$ & $5.7610\mathrm{E}{-}59$ & $6.8598\mathrm{E}{-}18$ & $4.5465\mathrm{E}{+}05$ & $9.7919\mathrm{E}{+}06$ & $3.0677\mathrm{E}{+}02$ & $6.5021\mathrm{E}{-}15$ \\
Zakharov & Ave & $\mathbf{0.0000E{+}00}$ & $3.7333\mathrm{E}{+}02$ & $1.2730\mathrm{E}{-}04$ & $1.9713\mathrm{E}{+}02$ & $6.0085\mathrm{E}{+}01$ & $1.7135\mathrm{E}{+}01$ & $1.6916\mathrm{E}{+}00$ \\
 & Std & $\mathbf{0.0000E{+}00}$ & $1.8380\mathrm{E}{+}02$ & $3.0267\mathrm{E}{-}04$ & $1.8984\mathrm{E}{+}02$ & $1.5209\mathrm{E}{+}01$ & $1.1586\mathrm{E}{+}01$ & $9.2641\mathrm{E}{-}01$ \\
Rosenbrock & Ave & $\mathbf{0.0000E{+}00}$ & $4.4404\mathrm{E}{+}00$ & $2.7428\mathrm{E}{+}01$ & $5.6324\mathrm{E}{+}03$ & $1.6724\mathrm{E}{+}02$ & $2.8011\mathrm{E}{+}01$ & $2.5242\mathrm{E}{+}01$ \\
 & Std & $\mathbf{0.0000E{+}00}$ & $8.6796\mathrm{E}{+}00$ & $7.1828\mathrm{E}{-}01$ & $1.7537\mathrm{E}{+}04$ & $1.0922\mathrm{E}{+}02$ & $2.6358\mathrm{E}{-}01$ & $6.7796\mathrm{E}{-}01$ \\
Rastrigin & Ave & $\mathbf{0.0000E{+}00}$ & $\mathbf{0.0000E{+}00}$ & $3.9790\mathrm{E}{-}14$ & $6.6335\mathrm{E}{+}01$ & $1.5836\mathrm{E}{+}02$ & $4.0773\mathrm{E}{+}00$ & $3.5344\mathrm{E}{+}01$ \\
 & Std & $\mathbf{0.0000E{+}00}$ & $\mathbf{0.0000E{+}00}$ & $6.0217\mathrm{E}{-}14$ & $2.3163\mathrm{E}{+}01$ & $2.0251\mathrm{E}{+}01$ & $1.2893\mathrm{E}{+}01$ & $4.2048\mathrm{E}{+}01$ \\
Ackley & Ave & $\mathbf{4.4409E{-}16}$ & $5.7732\mathrm{E}{-}15$ & $5.0386\mathrm{E}{-}13$ & $2.0995\mathrm{E}{+}00$ & $9.9140\mathrm{E}{-}01$ & $3.6543\mathrm{E}{-}03$ & $1.2893\mathrm{E}{-}11$ \\
 & Std & $\mathbf{0.0000E{+}00}$ & $3.8374\mathrm{E}{-}15$ & $2.7953\mathrm{E}{-}13$ & $7.3911\mathrm{E}{-}01$ & $7.3239\mathrm{E}{-}01$ & $1.8950\mathrm{E}{-}03$ & $2.3999\mathrm{E}{-}11$ \\
Griewank & Ave & $\mathbf{0.0000E{+}00}$ & $\mathbf{0.0000E{+}00}$ & $2.3380\mathrm{E}{-}03$ & $2.6057\mathrm{E}{-}01$ & $2.2497\mathrm{E}{-}01$ & $9.3142\mathrm{E}{-}03$ & $2.3444\mathrm{E}{-}03$ \\
 & Std & $\mathbf{0.0000E{+}00}$ & $\mathbf{0.0000E{+}00}$ & $4.9592\mathrm{E}{-}03$ & $2.7852\mathrm{E}{-}01$ & $1.4000\mathrm{E}{-}01$ & $2.7369\mathrm{E}{-}02$ & $7.4138\mathrm{E}{-}03$ \\
Schwefel & Ave & $\mathbf{3.8183E{-}04}$ & $1.1437\mathrm{E}{+}02$ & $9.4872\mathrm{E}{+}03$ & $4.2627\mathrm{E}{+}03$ & $6.4047\mathrm{E}{+}03$ & $5.7787\mathrm{E}{+}02$ & $3.4676\mathrm{E}{+}03$ \\
 & Std & $\mathbf{0.0000E{+}00}$ & $3.0375\mathrm{E}{+}02$ & $2.0807\mathrm{E}{+}02$ & $6.0223\mathrm{E}{+}02$ & $1.0335\mathrm{E}{+}03$ & $1.2097\mathrm{E}{+}03$ & $1.2593\mathrm{E}{+}03$ \\
Levy & Ave & $\mathbf{1.4998E{-}32}$ & $2.9172\mathrm{E}{-}02$ & $1.5013\mathrm{E}{+}00$ & $2.4669\mathrm{E}{+}00$ & $1.3304\mathrm{E}{+}00$ & $7.7895\mathrm{E}{-}03$ & $5.2315\mathrm{E}{+}00$ \\
 & Std & $\mathbf{0.0000E{+}00}$ & $5.7032\mathrm{E}{-}02$ & $1.1917\mathrm{E}{-}01$ & $3.2271\mathrm{E}{+}00$ & $9.3951\mathrm{E}{-}01$ & $3.7897\mathrm{E}{-}03$ & $6.8568\mathrm{E}{+}00$ \\
Michalewicz & Ave & $\mathbf{-2.2599E{+}01}$ & $-1.0123\mathrm{E}{+}01$ & $-8.3278\mathrm{E}{+}00$ & $-2.0990\mathrm{E}{+}01$ & $-1.1934\mathrm{E}{+}01$ & $-1.6928\mathrm{E}{+}01$ & $-1.9394\mathrm{E}{+}01$ \\
 & Std & $1.3254\mathrm{E}{+}00$ & $\mathbf{9.5161E{-}01}$ & $1.1664\mathrm{E}{+}00$ & $2.0907\mathrm{E}{+}00$ & $1.1312\mathrm{E}{+}00$ & $2.1227\mathrm{E}{+}00$ & $1.7703\mathrm{E}{+}00$ \\
\bottomrule
\end{tabular}%
}
\end{table*}

\begin{table}[t]
\centering
\caption{Exact-tie average ranks over ten benchmark functions.}
\label{tab:ranks}
\begin{tabular}{lcc}
\toprule
Method & Avg. rank & Best/tied \\
\midrule
WASHH & $\mathbf{1.10}$ & $\mathbf{10}$ \\
WOA & 2.90 & 2 \\
GWO & 4.00 & 0 \\
RandomHH & 4.10 & 0 \\
LWOA & 4.20 & 0 \\
PSO & 5.80 & 0 \\
DE & 5.90 & 0 \\
\bottomrule
\end{tabular}
\end{table}

\subsection{Benchmark Results}

Table~\ref{tab:results} reports a consistent advantage for WASHH across the benchmark suite.
WASHH is best or tied best on all ten functions and strictly improves over WOA on Sphere, Bent Cigar, Zakharov, Rosenbrock, Ackley, Schwefel, Levy, and Michalewicz.
On Rastrigin and Griewank, both WASHH and WOA reach a mean objective value of zero, so the comparison is treated as an exact tie rather than a strict improvement.
The same pattern appears in the average ranks: WASHH obtains rank 1.10, whereas the closest competitor, WOA, obtains rank 2.90.

The strongest gains occur on landscapes where a single WOA-style trajectory is insufficient.
WASHH reduces the mean objective on Zakharov from $3.73\times10^2$ to $0.00$ and on Rosenbrock from $4.44\times10^0$ to $0.00$.
It also reaches $3.82\times10^{-4}$ on Schwefel and $1.50\times10^{-32}$ on Levy, both below the corresponding WOA means.
On Michalewicz, WASHH improves over PSO, which is the strongest single-method baseline for that function.
These results support the central claim that combining adaptive behavior selection with anchor-guided refinement is more robust than relying on one fixed population update.

\begin{figure*}[t]
\centering
\includegraphics[width=\textwidth]{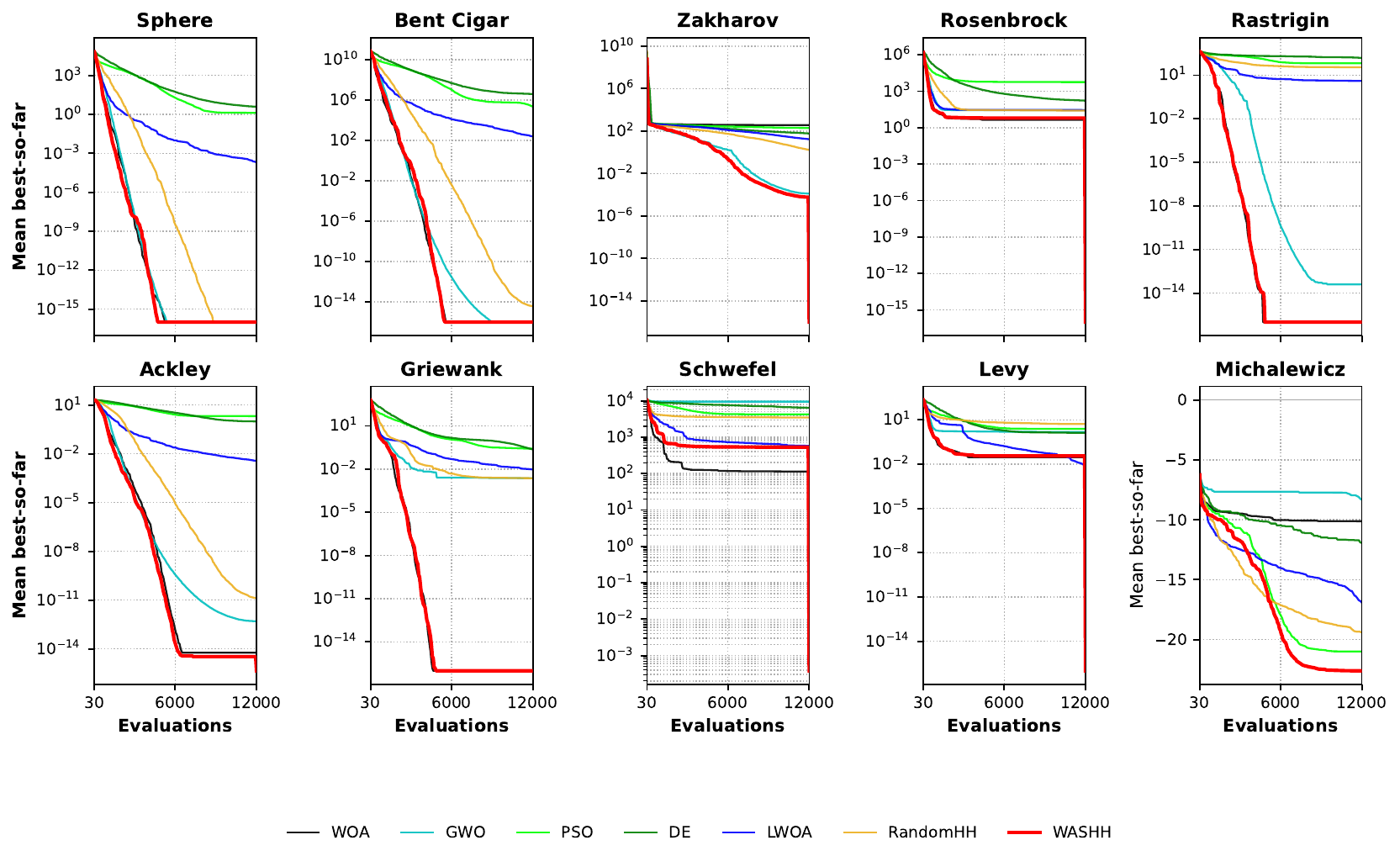}
\caption{Mean best-so-far convergence curves over 10 independent runs for all compared algorithms on the ten benchmark functions.}
\label{fig:convergence}
\end{figure*}

Figure~\ref{fig:convergence} visualizes the optimization process using mean best-so-far traces over the same 10 runs.
The curves show that WASHH reaches the best or tied-best final mean on every displayed function, and that the reserved refinement phase is especially useful when the domain-derived anchor neighborhoods are informative, as on Rosenbrock, Schwefel, and Levy.

\subsection{Ablation Study}

We further conduct a strict leave-one-component-out ablation for the two core components.
Each ablated variant starts from the same WASHH implementation and removes only one mechanism, and all variants share the same problem and run seeds.
Using common seeds makes the comparison focus on the removed component rather than ordinary run-to-run variation.

\begin{table}[t]
\centering
\caption{Ablation results over ten benchmark functions.}
\label{tab:ablation}
\resizebox{\columnwidth}{!}{%
\begin{tabular}{lcc}
\toprule
Variant & Avg. rank & Best/tied \\
\midrule
Full WASHH & $\mathbf{2.10}$ & $\mathbf{10}$ \\
w/o adaptive selection & 2.30 & 9 \\
w/o anchor refinement & 3.50 & 3 \\
\bottomrule
\end{tabular}%
}
\end{table}

Table~\ref{tab:ablation} identifies anchor refinement as the most influential component on this benchmark suite.
When anchors are removed, the best/tied count drops from 10 to 3 and the average rank increases from 2.10 to 3.50, with clear degradation on Zakharov, Rosenbrock, Schwefel, and Levy.
Removing adaptive selection has a smaller effect on anchor-friendly functions, but it degrades Michalewicz, where the full method benefits from selecting a memory-based behavior rather than following the default WOA-guided path.

The exact-tie ranking avoids overstating improvements on functions where strong methods already reach the numerical optimum.
For Rastrigin and Griewank, WASHH is counted as tied rather than superior.
The practical advantage comes from the remaining functions, where different baselines reveal different weaknesses: WOA converges slowly on Zakharov, LWOA is less stable on Levy, and PSO or DE can miss the final refinement required to reach very small objective values.
This behavior matches the intended LEAD use case, in which the optimizer should reuse multiple search behaviors and adapt their use to the task rather than assuming that one population rule is always appropriate.

\section{Breast Cancer Diagnosis Model Configuration}

We use breast cancer diagnosis model configuration as an application scenario.
The task uses the Wisconsin Diagnostic Breast Cancer dataset distributed with scikit-learn~\cite{pedregosa2011scikit,street1993nuclear}.
The classifier is an RBF-kernel support vector classifier.
The optimizer configures two continuous hyperparameters, $C$ and $\gamma$, encoded in log scale.
The objective is validation log loss on a stratified holdout split, so each black-box evaluation trains an SVC model and evaluates probabilistic predictions.

The case study fits the LEAD setting because reference configurations are available before optimization begins but remain insufficient without validation on the target split.
Default SVC settings and common experience-based choices provide low-cost anchors, whereas the final validation loss must still be measured by training and evaluating the model.
WASHH uses these anchors only for refinement; all compared optimizers receive the same budget of 300 model evaluations, the same population size of 30, and 10 independent optimizer seeds.

\begin{table}[t]
\centering
\caption{Breast cancer diagnosis HPO validation log loss over 10 runs.}
\label{tab:breast}
\begin{tabular}{lcc}
\toprule
Method & Ave $\downarrow$ & Std $\downarrow$ \\
\midrule
\textbf{WASHH} & $\mathbf{0.076273}$ & $\mathbf{0.000000}$ \\
PSO & 0.076361 & 0.000087 \\
RandomHH & 0.076460 & 0.000210 \\
GWO & 0.076510 & 0.000394 \\
DE & 0.076573 & 0.000331 \\
LWOA & 0.076895 & 0.000786 \\
WOA & 0.077961 & 0.001541 \\
\bottomrule
\end{tabular}
\end{table}

Table~\ref{tab:breast} shows that WASHH obtains the lowest mean validation log loss.
The best configuration found by WASHH is approximately $C=5.21$ and $\gamma=0.0105$.
The improvement over PSO is small in magnitude but stable across the 10 seeds, while the improvement over WOA is larger.
This result illustrates the role of anchor refinement: the method can exploit domain-default regions while still using adaptive population search to refine the final configuration.

\begin{figure}[t]
\centering
\includegraphics[width=\columnwidth]{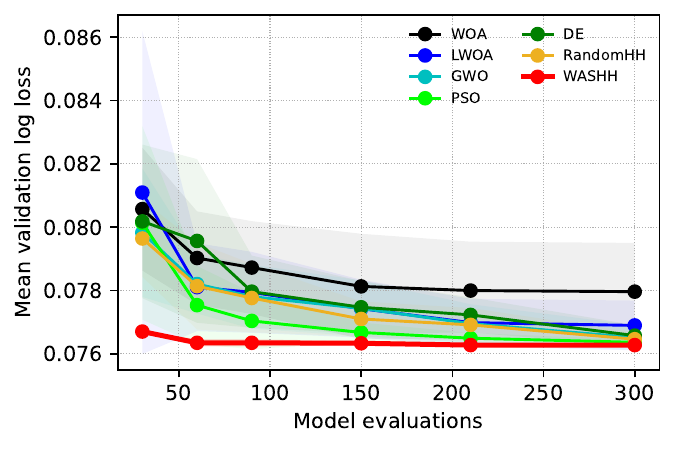}
\caption{Breast cancer diagnosis model configuration. The curves show mean validation log loss during the HPO search, with shaded one-standard-deviation bands over 10 runs.}
\label{fig:breast}
\end{figure}

Figure~\ref{fig:breast} shows the model-configuration process for this case study.
Because the classifier is an SVC rather than a neural network, the relevant optimization trace is not an epoch-by-epoch training curve.
Instead, each point corresponds to a completed model evaluation with a candidate $(C,\gamma)$ pair, and the curve shows how validation log loss decreases across the HPO budget.

\begin{figure}[t]
\centering
\includegraphics[width=\columnwidth]{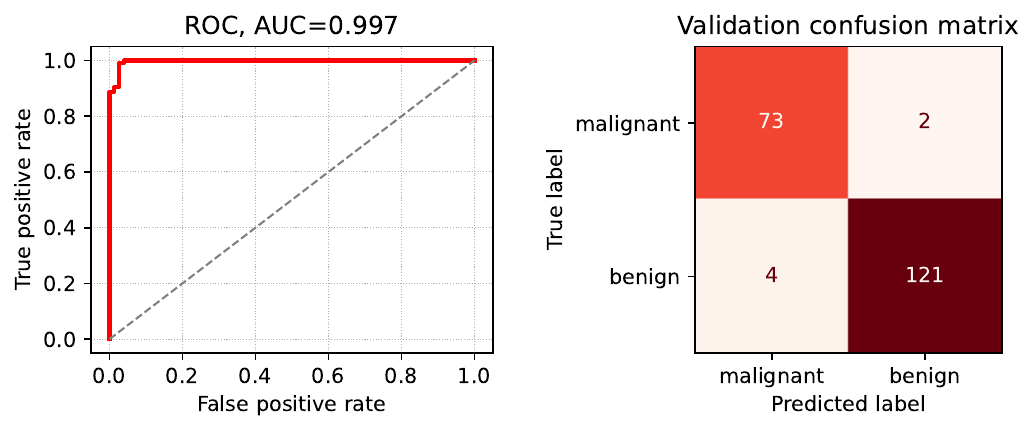}
\caption{Final validation-set diagnostic performance of the SVC configured by WASHH, shown by the ROC curve and confusion matrix.}
\label{fig:breastdiag}
\end{figure}

Figure~\ref{fig:breastdiag} reports the final diagnostic behavior of the configured classifier.
It complements the HPO trace by showing the validation-set discrimination and error pattern after the selected configuration has been fitted.

\section{Discussion}

The empirical results point to the value of combining adaptive behavior selection with task-level reference information.
WASHH does not rely on a single update rule being uniformly superior.
Instead, the controller maintains a small portfolio of complementary search behaviors, while anchor refinement uses low-cost reference configurations to intensify search in regions that are plausible but still require black-box validation.
This design is well matched to algorithm and model configuration, where default parameters, box centers, or historically effective settings are often available before optimization begins.

The ablation results further clarify the mechanism.
Anchor refinement accounts for most of the measured gain on the deterministic benchmark suite, especially on functions where useful reference neighborhoods exist.
Adaptive selection contributes a smaller but still visible benefit by allowing the search to move away from the default WOA-guided behavior when another update pattern is more suitable.
The method should therefore be interpreted as an anchor-aware selection hyper-heuristic rather than as a universally dominant population update.

The current evaluation is intentionally controlled.
It covers ten deterministic continuous benchmarks and a compact breast cancer diagnosis HPO case, which is sufficient to test the proposed mechanism under fixed budgets but not sufficient to characterize all deployment conditions.
Future work should evaluate WASHH on noisy and constrained objectives, rotated or shifted landscapes, larger hyperparameter spaces, and real applications where each objective evaluation is substantially more expensive.

\section{Conclusion}

This paper introduced WASHH, a Whale-guided Adaptive Selection Hyper-Heuristic for budgeted continuous black-box optimization.
The method frames WOA-guided exploitation, memory-based search, multi-leader guidance, differential variation, local refinement, and anchor refinement as selectable behaviors within one online controller.
Across ten 30-dimensional benchmark functions, WASHH achieves the best average rank, is best or tied best on every function, and strictly improves over WOA on eight functions.
The ablation study shows that anchor refinement is the dominant measured contributor on this suite, with adaptive behavior selection providing additional robustness.
In the breast cancer diagnosis HPO case, WASHH also obtains the lowest mean validation log loss under the same model-evaluation budget.
These findings indicate that lightweight, anchor-aware selection hyper-heuristics are a useful direction for LEAD when prior configurations are available but still need task-specific optimization.

\section*{Acknowledgments}

This work was supported by Macao Polytechnic University (RP/FCA-01/2025) and the Macao Science and Technology Development Fund (FDCT-MOST: 0018/2025/AMJ).
This support enabled data collection, analysis, and interpretation, and covered expenses related to research materials and participant recruitment.

\end{document}